# Improvement of PSO algorithm by memory based gradient search - application in inventory management

**Tamás Varga, András Király, János Abonyi**

University of Pannonia, Department of Process Engineering,

10. Egyetem Str., H-8200, Veszprém, Hungary

e-mail vargat@fmt.uni-pannon.hu

*Abstract* — Advanced inventory management in complex supply chains requires effective and robust nonlinear optimization due to the stochastic nature of supply and demand variations. Application of estimated gradients can boost up the convergence of Particle Swarm Optimization (PSO) algorithm but classical gradient calculation cannot be applied to stochastic and uncertain systems. In these situations Monte-Carlo (MC) simulation can be applied to determine the gradient. We developed a memory based algorithm where instead of generating and evaluating new simulated samples the stored and shared former function evaluations of the particles are sampled to estimate the gradients by local weighted least squares regression. The performance of the resulted regional gradient-based PSO is verified by several benchmark problems and in a complex application example where optimal reorder points of a supply chain are determined.

*Keywords* — particle swarm optimization, gradient, Monte-Carlo simulation, inventory management system

I. Introduction

In the last decades, optimization was featured in almost all aspects of human civilization, thus it has truly become an indispensable method. In some aspects, even a local optima can highly improve the efficiency or reduce the expenses, however, most companies want to keep their operational costs as low as possible, i.e. on global minimum. Problems where solutions must satisfy a set of constraints are known as constrained optimization problems. In inventory control theory, one of the most important and most strict constraints is the service level, i.e. the portion of satisfied demands from all customer needs (Schwartz, Wang & Rivera 2006). Stochastic nature of the supply and demand variations in complex supply chains require effective and robust nonlinear optimization for advanced inventory management.

There are two popular swarm inspired methods in computational intelligence areas: Ant Colony Optimization (ACO) and Particle Swarm Optimization (PSO). ACO was inspired by the behaviors of ants and has many successful applications in discrete optimization problems. The particle swarm concept originated as a simulation of simplified social system. Next to these two there can be found many other swarm intelligence based method in literature to solve optimization problems, such as Firefly Algorithm (FA); Bat Algorithm (BA); Krill Herd Algorithm (KHA). FAs are recently developed methods to optimize nonlinear design problems based on the idealized behavior of the flashing characteristics of fireflies (Yang 2010a). BA is very similar to the PSO however it is based on the hunting method of bats using their echolocation ability (Yang 2010b). One of the latest algorithms is the KHA wherein the benefits of swarm intelligence and the genetic algorithms are integrated which results a reliable optimization technique with good conversion rate (Gandomi & Alavi 2012).

Particle swarm model can be used to solve stochastic and constrained optimization problems (Hu & Eberhart 2002, Kennedy & Eberhart 1995). The particle swarm concept originated as a simulation of simplified social system. In PSO, the potential solutions, called particles, fly through the problem space by following the current optimum particles. All of particles have fitness values which are evaluated by the fitness function to be optimized, and have velocities in the direction based on their inertia, best fitness value and the best solution found by the population. PSO is getting more and more widespread tool in solving complex engineering problems since it is easily interpretable and implementable optimization algorithm, and it can be effectively applied to find the extremum of nonlinear optimization problems with many independent parameters. The particle swarm optimization algorithm has been successfully applied to a wide set of complex problems, like data mining (Sousaa, Silvaa & Neves 2004), software testing (Windisch, Wappler & Wegener 2007), nonlinear mapping(Edwards, Engelbrecht & Franken 2005), function minimization (Kennedy & Eberhart 1995) or neural network training (Engelbrecht, Engelbrecht & Ismail 1999) and in the last decade, constrained optimization using PSO got a bigger attention (Hu & Eberhart 2002, Parsopoulos & Vrahatis 2002, Wimalajeewa & Jayaweera 2008).

There exist some well-known conditions under which the basic PSO algorithm exhibits poor convergence characteristics (Bergh 2002). However, only a few studies have considered the hybridization of PSO, especially making use of gradient information directly within PSO. Notable ones are HGPSO (Noel & Jannett 2004) and GTPSO (Zhang, Zhang & Zhang 2009), which use the gradient descent algorithm, and FR-PSO (Borowska & Nadolski 2009), which applies the Flecher-Reeves method. As it will be demonstrated in the following sections, combining these two methods appropriately, the efficiency of the optimization using PSO can be considerably improved.

Classical gradient calculation cannot be applied to stochastic and uncertain systems. In these situations stochastic techniques like Monte-Carlo (MC) simulation can be applied to determine the gradient. These techniques require additional function evaluations. We developed a more economic, memory based algorithm where instead of generating and evaluating new simulated samples the stored and shared former function evaluations of the particles are sampled to estimate the gradients by local weighted least squares regression. The performance of the resulted fully informed, regional gradient based PSO is verified by several benchmark problems.

The algorithm has been applied to find the optimal reorder points of a supply chain. The stochastic objective function is based on the linear combination of holding cost in the warehouses, the order cost and the unit price. The inequality constraints are defined based on the minimal service level values. The determination of safety stock in an inventory model is one of the key tasks of supply chain management. Miranda and Garrido include safety stock in the inventory model in (Miranda & Garrido 2004). Authors in (Graves & Willems 2008) give a model for positioning safety stock in a supply chain subject to non-stationary demand and show how to extend their former model to find the optimal placement safety stocks under constant service time (CST) policy. Prékopa in (Prékopa 2006) gives an improved model for the so called Hungarian inventory control model to find the minimal safety stock level that ensures the continuous production, without disruption. The bullwhip effect is an important phenomenon in supply chains. Authors in (Makajic-Nikolic, Panic & Vujoševic 2004) show how a supply chain can be modeled and analyzed by colored petri nets (CPN) and CPN tools and they evaluate the bullwhip effect, the surplus of inventory goods, etc. using the beer game as demonstration. More recent research can be found in (Caloieroa, Strozzia & Comenges 2008), which shows that an order policy applied to a serial single-product supply chain with four echelons can reduce or amplify the bullwhip effect and inventory oscillation. Miranda et al. investigate the modeling of a two echelon supply chain system and optimization in two steps (Miranda & Garrido 2009), while a massive multi-echelon inventory model is presented by Seo (Seo 2006), where an order risk policy for general multi-echelon system is given, which minimizes the system operation cost. A really complex system is examined in (Srinivasan & Moon 1999), where it is necessary to apply some clustering for similar items, because detailed analysis could become impossible considering each item individually. The stability of the supply chain is also an intensively studied area. (Nagatania & Helbing 2004) shows that a linear supply chain can be stabilized by the anticipation of the own future inventory and by taking into account the inventories of other suppliers, and Vaughan in (Vaughan 2006) presents a linear order point/lot size model that with its robustness can contribute to business process modeling.

We developed a Monte-Carlo simulator which uses probability distributions based on material usage data posted in the logistic module of an enterprise resource planning (ERP) system. The main objective of this development was to build a simulator that can use simple building blocks to construct models of complex supply chain networks. With the synergistic combination of this tool and the proposed PSO algorithm we minimized the inventory holding cost by changing the parameters of our operational space while keeping the service level at the required value. The results illustrate the benefits of the incorporation of the regional gradients into the PSO algorithm.

## II. THE IMPROVED PSO ALGORITHM

### A. Classical PSO algorithm

The original intent was to graphically simulate the choreography of bird of a bird block or fish school. However, it was found that particle swarm model can be used as an optimizer. Suppose the following scenario: a group of birds are randomly searching food in an area. There is only one piece of food in the area being searched. All the birds do not know where the food is. But they know how far the food is in each iteration. So what's the best strategy to find the food? The effective one is to follow the bird which is nearest to the food. PSO is based on this scheme. This stochastic optimization technique has been developed by Eberhart and Kennedy in 1995 (Kennedy & Eberhart 1995). In PSO, the potential solutions, called particles, fly through the problem space by following the current optimum particles. All of particles have fitness values which are evaluated by the fitness function to be optimized, and have velocities which direct to the flying of the particles.

PSO is initialized with a group of random particles (solutions) and then searches for optima by updating generations. In every iteration, each particle is updated by following two "best" values. The first one is the best solution (fitness) it has achieved so far. (The fitness value is also stored.) This value is called pbest. Another "best" value that is tracked by the particle swarm optimizer is the best value, obtained so far by any particle in the population. This best value is a global best and called *gbest*. When a particle takes part of the population as its topological neighbors, the best value is a local best and is called *lbest*.

$$\mathbf{v}_j(k+1) = w \cdot \mathbf{v}_j(k) + c_1 \cdot rand() \cdot (\mathbf{x}_{pbest,j} - \mathbf{x}_j(k)) + c_2 \cdot rand() \cdot (\mathbf{x}_{gbest} - \mathbf{x}_j(k)) \quad (1)$$

$$\mathbf{x}_j(k+1) = \mathbf{x}_j(k) + \mathbf{v}_j(k+1) \cdot dt \quad (2)$$

where $j = 1,...,\lambda$ represents the index of the *j*th swarm, **v** is the particle velocity, *rand()* is a random number between [0,1], $c_1$, $c_2$ are learning factors. Code 1. shows the pseudo code of the classical PSO algorithm.

Code 1.: The pseudo code of the PSO algorithm

```
procedure PSO; {
        Initialize particles;
        while (not terminate) do {
                for each particle {
                        Calculate fitness value;
                        if fitness <pBest than pBest = fitness;
                }
                Choose the best particle as the gBest;
                for each particle {
                        Calculate particle velocity;
                        Update particle position;
                }
        }
}
```

The role of the, *w*, inertia weight in Eq. (1), is considered critical for the convergence behavior of PSO. The inertia weight is employed to control the impact of the previous history of velocities on the current one. Accordingly, the parameter regulates the trade–off between the global and local exploration abilities of the swarm. A large inertia weight facilitates global exploration (searching new areas) while a small one tends to facilitate local exploration, i.e. fine–tuning the current search area.

PSO shares many similarities with evolutionary computation techniques, e.g. with evolutionary algorithms (EAs). Both algorithms start with a group of a randomly generated population, both have fitness values to evaluate the population. Both update the population and search for the optimum with random techniques. Both systems do not guarantee success. The main difference between these algorithms is that PSO does not have genetic operators like crossover and mutation. Particles update themselves with the internal velocity. They also have memory, which is important to the algorithm.

Compared with evolutionary algorithms, the information sharing mechanism in PSO is significantly different. In EAs, chromosomes share information with each other. So the whole population moves like a one group towards an optimal area. In PSO, only *gBest* (or *lBest*) gives out the information to others. It is a one-way information sharing mechanism, the evolution only looks for the best solution. Compared with EAs, all the particles tend to converge to the best solution quickly even in the local version in most cases. Compared to EA, the advantages of PSO are that PSO is easy to implement and there are few parameters to adjust. Hence, PSO has been successfully applied in many areas: function optimization, artificial neural network training (Engelbrecht, Engelbrecht & Ismail 1999), control (Victoirea & Jeyakumar 2004), scheduling (Wimalajeewa & Jayaweera 2008), and other areas where GA can be applied.

The basic PSO algorithm exhibits poor convergence characteristics under some specific conditions. We gave a small overview also about the previous gradient based methods, and in this section we will demonstrate a novel way, how the particle swarm optimization (PSO) technique can be improved with the calculation of the gradient of the applied objective function. There are some well documented algorithms in the literature to boost the convergence of the basic PSO algorithm. Victoirea et al. developed a hybrid PSO to solve the economic dispatch program (Victoirea & Jeyakumar 2004). They combined PSO with Sequential Quadratic Programming to search for the gradient of the objective function. A very similar algorithm is introduced by Noel, in which quasi Newton-Raphson (QNR) algorithm is applied to calculate the gradient (Noel & Jannett 2004). The QNR algorithm optimizes by locally fitting a quadratic surface and finding the minimum of that quadratic surface.

### B. *Improved PSO algorithm*

Our aim is to develop a novel PSO algorithm which is able to consider linear and non-linear constraints and it calculates the gradient of the objective function to improve the affectivity. PSO is initialized with a group of random particles (solutions) and then searches for optima by updating generations. In every generation, each particle is updated by following two "best" values. The first one is the best solution (fitness) it has achieved so far. This value is called *pbest*. Another "best" value that is tracked by the particle swarm optimizer is the best value, obtained so far by any particle in the population. This best value is a global best and called *gbest*. When a particle takes part of the population as its topological neighbors, the best value is a local best and is called *lbest*. Our vision is to apply the gradient of the objective function in every generation to control the movements of the particles. Therefore, the equation which is applied to calculate the velocity of the particles is modified:

$$\mathbf{v}_j(k+1) = w \cdot \mathbf{v}_j(k) + c_1 \cdot rand() \cdot (\mathbf{x}_{pbest,j} - \mathbf{x}_j(k)) + c_2 \cdot rand() \cdot (\mathbf{x}_{gbest} - \mathbf{x}_j(k)) + c_3 \cdot \mathbf{g}_j(f(\mathbf{x}(k))) \quad (3)$$

where $\mathbf{g}_j(f(\mathbf{x}(k))) = \dfrac{\partial f(\mathbf{x}(k))}{\partial \mathbf{x}_j(k)}$ represents the partial derivatives of the objective function, and $c_3$ is the weight for the gradient term.

It should be noted that this concept can be interpreted as inserting a gradient-descent update step into the iterations of classical PSO, $\mathbf{x}(k+1) = \mathbf{x}(k) - \eta \nabla f(\mathbf{x}(k))$ where the learning rate is equal to $\eta = c_3 dt$.

The above algorithm can be applied only to continuously differentiable objective functions $\nabla f(\mathbf{x}(k))$. The simples approach to calculate the gradient is HGPSO (Noel & Jannett 2004) is the numerical approximation of the gradient.

$$\frac{\partial}{\partial x_i}(\mathbf{x}(k)) = \frac{f(\mathbf{x}(k) + E_i \varepsilon) - f(\mathbf{x}(k))}{\varepsilon} \quad (4)$$

The main drawback of this approach is that the $\varepsilon$ step size is difficult to design and the whole approach is selective to noise and uncertainties. It is interesting to note that PSO itself can also be interpreted as a gradient based search algorithm where point differences are used as approximation of the regional gradient. The normalized gradient evaluated as the point difference method is

$$\mathbf{e} = \frac{f(\mathbf{x}_j) - f(\mathbf{x}_i)}{\|(\mathbf{x}_j - \mathbf{x}_i)\|} \quad (5)$$

This point–difference estimate can be considered as regional gradient for the local region of $\mathbf{x}_i$ and $\mathbf{x}_j$. Hence, the velocity of PSO can be interpreted as a weighted combination of a point-difference estimated global regional gradient $(x_{gbest} - x_j(k))$ and a point-difference estimated finer regional gradient $(x_{pbest} - x_j(k))$.

Our aim is to further improve the optimization by providing robust yet accurate estimation of gradients. To obtain a robust estimate a so called regional gradient should be calculated. When the function is differentiable the gradient for a region $S$ it is calculated as

$$\nabla f(\mathbf{x})^* = \frac{1}{volume(S)} \int_{x \in S} \nabla f(\mathbf{x}) d\mathbf{x} \qquad (6)$$

where $S$ represents the local region where the gradient is calculated.

However, when heuristic optimization algorithm should be applied the objective function is mostly not continuously differentiable or not explicitly given due to limited knowledge. Therefore the gradient is calculated as

$$\nabla f(\mathbf{x})^* = \frac{\int_{x \in S} f(\mathbf{x}) d\mathbf{x}}{\int_{x \in S} d\mathbf{x}} \qquad (7)$$

An interesting example for this approach is how regional gradient is calculated in the Evolutionary-Gradient-Search (EGS) procedure proposed by R. Solomon (Salomon & Arnold 2009). In EGS at each iteration generates $\lambda$ test candidates by applying random "mutations" of $\mathbf{x}(k)$.

$$\mathbf{v}_i = \mathbf{x}(k) + \mathbf{z}_i \qquad (8)$$

where $\mathbf{z}_i$ is a Gaussian distributed variable with zero mean and standard deviation $\sigma/\sqrt{n}$. For $n \gg 1$ these test points will be distributed on a hypersphere with radius $\sigma$. By using information given by all candidates the procedure calculates the gradient and a unit vector $\mathbf{e}(k)$ that points into the direction of the estimated gradient:

$$\mathbf{g}(k) = \sum_{i=1}^{\lambda} f(\mathbf{v}_i) - f(\mathbf{x}(k))(\mathbf{v}_i - \mathbf{x}(k)) \qquad (9)$$

$$\mathbf{e}(k) = \frac{\mathbf{g}(k)}{\|\mathbf{g}(k)\|} \qquad (10)$$

These techniques require additional function evaluations. It is important to note that this concept discards all information related to the evaluation of $\mathbf{v}_i$.

We developed a more economic, memory based algorithm where instead of generating and evaluating new simulated samples the stored and shared former function evaluations of the particles are sampled to estimate the gradients by local weighted least squares regression.

This idea is partly similar to the concept of the fully-informed particle swarm (FIPS) algorithm proposed by Mendes (Mendes, Kennedy & Neves 2004). FIPS that can be also considered as a hybrid method for estimating the gradient by a point difference of the weighted regional gradient estimate $(P_j - \mathbf{x}(k))$ based on the *lbest* solutions and adding an additional gradient related term to the velocity adaptation:

$$v_j(k+1) = \ldots + c_3\left(v_j(k-1) + \varphi\left(P_j(k) - \mathbf{x}(k)\right)\right) \tag{11}$$

$$P_j(k) = \frac{\sum_{k \in S} \varphi_k \mathbf{x}_{lbest,i}(k)}{\sum_{k \in S} \varphi_k} \tag{12}$$

where $\varphi_k$ is drawn independently from the uniform distribution.

FIPS utilizes only the current $\mathbf{x}_{lbest,i}(k)$ values so it does not have a memory.

The main concept of our work is the effective utilization of the previous function evaluations. So instead of generating new and new samples and loosing information from previous generations the whole trajectories of the particles are utilized.

The weighted regression problem that gives a robust estimates of the gradients is formulated by arranging these former function evaluations $\{\mathbf{x}(k), f(\mathbf{x}(k))\}$ into indexed data pairs $\{\mathbf{v}_i, f(\mathbf{v}_i)\}$ and calculating the following differences $\Delta f_i(k) = f(\mathbf{v}_i) - f(\mathbf{x}(k))$, $\Delta x_i(k) = \mathbf{v}_i - \mathbf{x}(k)$

$$\Delta \mathbf{f}(k) = \begin{bmatrix} \Delta f_1(1) \\ \Delta f_\lambda(1) \\ \Delta f_1(k-1) \\ \Delta f_\lambda(k-1) \end{bmatrix}, \Delta \mathbf{X}(k) = \begin{bmatrix} \Delta \mathbf{x}_1(1) \\ \Delta \mathbf{x}_\lambda(1) \\ \Delta \mathbf{x}_1(k-1) \\ \Delta \mathbf{x}_\lambda(k-1) \end{bmatrix}$$

where $\lambda$ represents the number of particles.

The weighted least squares estimate is calculated as

$$\mathbf{g}_j(k) = \left(\Delta \mathbf{X}^T(k)\mathbf{W}_j(k)\Delta \mathbf{X}(k)\right)^{-1} \Delta \mathbf{X}^T(k)\mathbf{W}_j^T(k)\Delta \mathbf{f}(k))$$ (13)

where the $\mathbf{W}_j(k)$ weighting matrix is a diagonal matrix representing the region of the $j^{th}$ particle. Similarly to EGS a Gaussian distributed weighting is used:

$$\beta_{j,i}(k) = \frac{1}{(2\pi)^{n/2}|\mathbf{\Sigma}|^{1/2}} \exp\left(\frac{1}{2}(\mathbf{v}_i - \mathbf{x}(k))^T \mathbf{\Sigma}^{-1}(\mathbf{v}_i - \mathbf{x}(k))\right)$$ (14)

where $\mathbf{v}_i$ the $i^{th}$ row of the $\Delta \mathbf{X}$ matrix, $j$ represents the $j^{th}$ particle, $w_{j,i}(k) = \dfrac{\beta_{j,i}(k)}{\sum_{i=1}\beta_{j,i}(k)}$ is the normalized probability of the sample, $\mathbf{\Sigma} = I\sigma$ is a diagonal matrix where $\sigma$ parameter represents the size of the region used to calculate the gradients. By using the information given by all previous states of the particles it is possible to calculate a unit vector that points into the direction of the estimated (global) gradient. The resulted algorithm is given in Code 2.

Code 2.: The pseudo code of the improved PSO algorithm

```
procedure improved PSO; {
    Initialize particles;
    while (not terminate) do {
        for each particle {
            Calculate fitness value;
            if fitness <pBest than pBest = fitness;
        }
        Choose the best particle as the gBest;
        for each particle {
            Calculate local gradient {
                Calculate normalized distance base
                weights of previous function evaluations particles by Eq. 14;
                Calculate regional gradients by Eq. 13;
            }
            Calculate particle velocity by Eq. 3;
            Update particle position by Eq. 2;
            Store particle position and related cost function in a database, $\{\mathbf{v}_i, f(\mathbf{v}_i)\}$
        }
    }
}
```

## C. Results

We tested the novel algorithm using several functions, including deterministic and stochastic ones as well. Figure 1 presents four of them and Table 1 contains the mathematical representation of the analyzed functions.

Table 1. Mathematical equations of the analyzed functions.

| Function | Equation |
|---|---|
| dropwave | $f(x,y) = -\dfrac{1 + \cos(12\sqrt{x^2 + y^2})}{\frac{1}{2}(x^2 + y^2) + 2}$ |
| griewangks | $f(x) = \dfrac{1}{4000}\sum_{i=1}^{n} x_i^2 - \prod_{i=1}^{n} \cos\left(\dfrac{x_i}{\sqrt{i}}\right) + 1$ |
| griewangks with noise | $f(x) = \dfrac{1}{4000}\sum_{i=1}^{n} x_i^2 - \prod_{i=1}^{n} \cos\left(\dfrac{x_i}{\sqrt{i}}\right) + 1 + rand()$ |
| stochastic function | $f(x,y) = \sin(120/x) + \cos(60/y)$ |

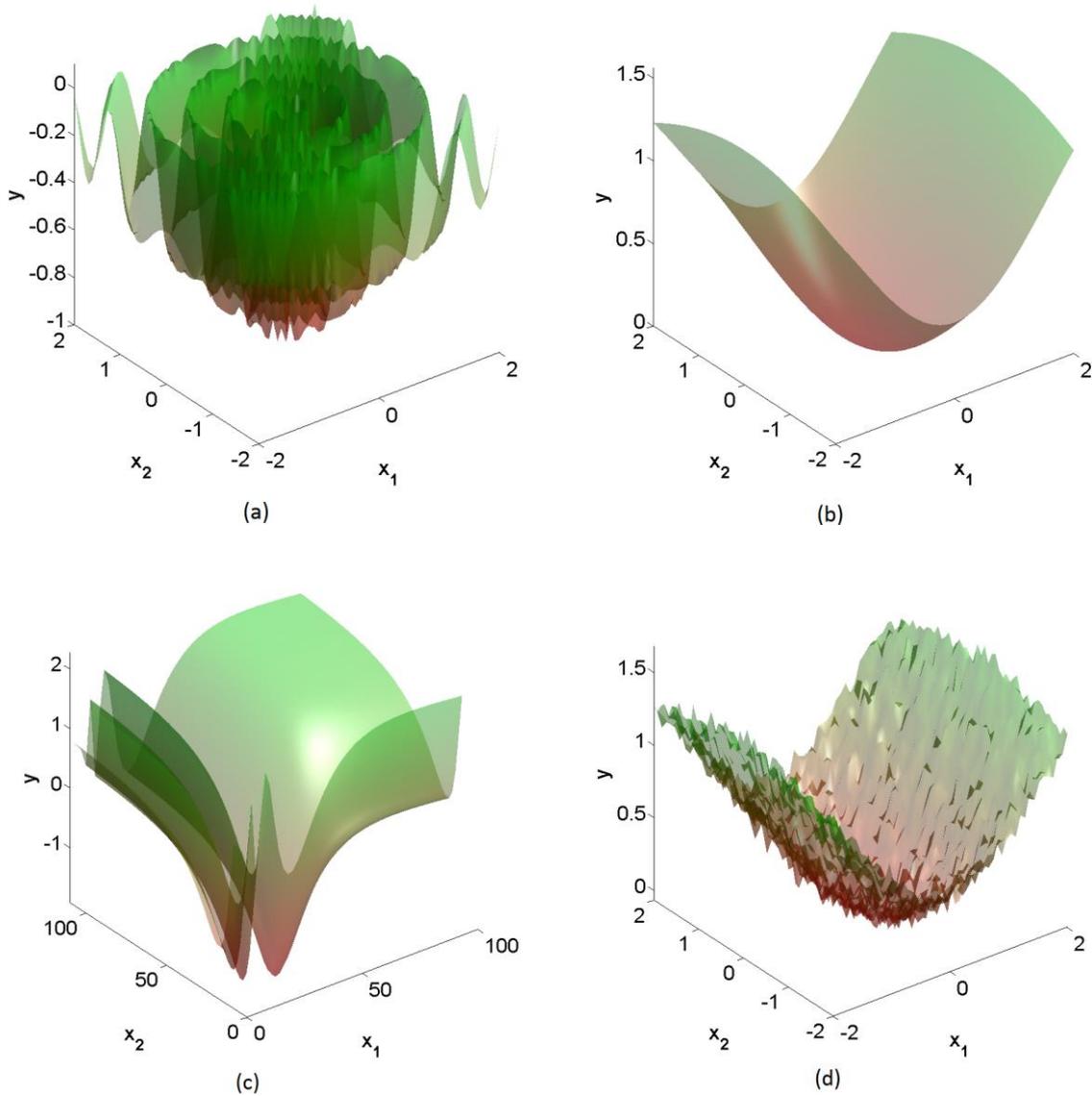

Figure 1 - Surface of the fitness function called "dropwave" (a) the "griewangks" function (b), the stochastic function we used (c), and a stochastic version of "griewangks" (d).

During our tests, the found global best value from the population, the mean of the best values of each individual, the iteration number what algorithm performed before termination and the iteration number when the global best was found were registered. In all tests, we applied Monte Carlo method to evaluate the performance of our approach effectively, thus the weight of the gradient part in the objective value calculation for the individuals was adjusted by 0.1 in the domain of [0, 1.25], and for each gradient weight, 500 MC simulations were performed. The result of our tests for the *g_best* values is presented in Figure 2. It can be deducted from the figure that the best global optima was found using 0.7 as the weight of the gradient part (since it is a minimization task), while the standard deviation is slightly smaller than in the classic PSO which is presented by the first data point in Figure 2., where the weight is equal to 0 (*w-grad = 0*). To prove the effectiveness of our method, we present the details of our tests in Table 2. The table presents our tests including two different deterministic functions (dropwave, griewanks) and two stochastic (griewankgs with noise and another stochastic function), highlighting the best results in each cases.

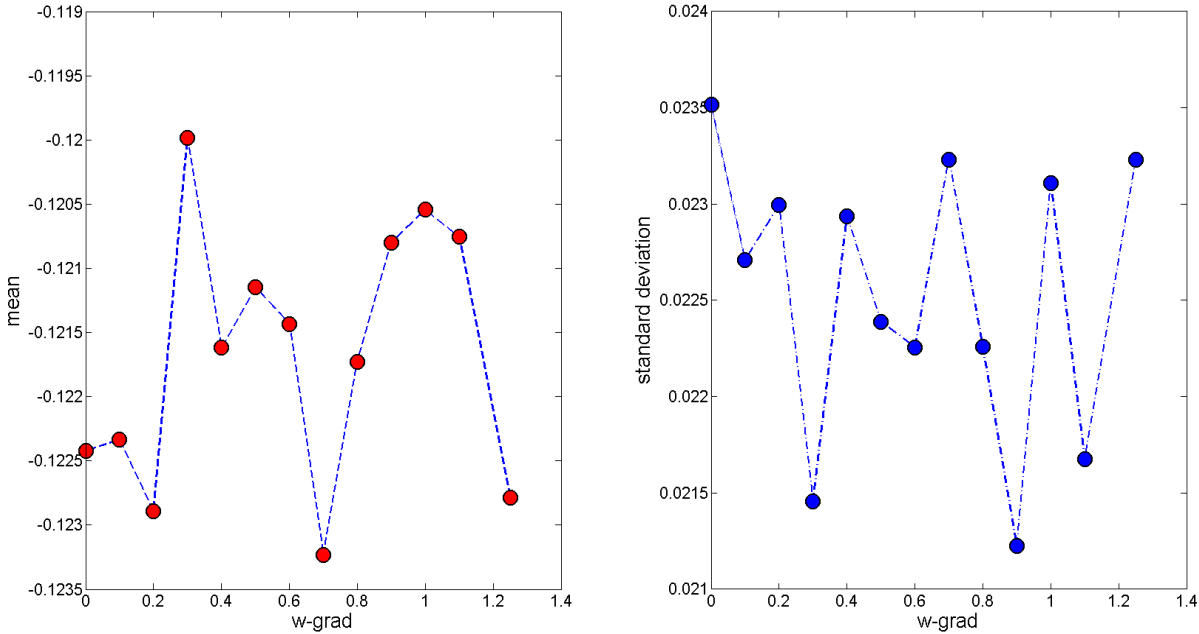

Figure 2: Histograms for the *"g_best"* values using the function called *"griewangks"* with noise. In the title of the subfigures, mean represents the mean value of the histogram, std is the standard deviation and *w-grad* is the weight of the gradient part in the objective value calculation of the individuals.

Table 2: Test results performing 500 MC simulation, modifying the weight for the gradient part. Best results are highlighted in each row for the objective values.

| Functions | Weights for the gradient | | | | | | | | | | | | |
|---|---|---|---|---|---|---|---|---|---|---|---|---|---|
| | 0 | 0.1 | 0.2 | 0.3 | 0.4 | 0.5 | 0.6 | 0.7 | 0.8 | 0.9 | 1.0 | 1.1 | 1.25 |
| Mean of g_best values | | | | | | | | | | | | | |
| dropwave $*10^{-1}$ | -9.98 | -9.99 | -9.99 | -9.99 | -9.98 | -9.96 | -9.94 | -9.92 | -9.87 | -9.77 | -9.65 | -9.50 | -9.40 |
| griewangks $*10^{-7}$ | 3.81 | 3.17 | 2.26 | 1.91 | 1.41 | 1.33 | 1.14 | 1.02 | 1.04 | 1.31 | 1.99 | 3.11 | 5.04 |
| griewangks with noise $*10^{-1}$ | -1.22 | -1.22 | -1.22 | -1.19 | -1.21 | -1.21 | -1.21 | -1.23 | -1.21 | -1.20 | -1.20 | -1.20 | -1.22 |
| Stochastic function | -9.56 | -9.58 | -9.58 | -9.590 | -9.596 | -9-591 | -9.586 | -9.582 | -9.56 | -9.55 | -9.54 | -9.52 | -9.47 |
| Mean of the mean of best values of each individual | | | | | | | | | | | | | |
| dropwave $*10^{-1}$ | -8.45 | -9.04 | -8.73 | -7.95 | -6.82 | -5.81 | -5.49 | -5.42 | -5.47 | -5.68 | -6.24 | -7.34 | -7.72 |
| griewangks $*10^{-4}$ | 226 | 134 | 71.6 | 40.5 | 19.9 | 10.5 | 5.66 | 3.23 | 1.78 | 1.24 | 1.28 | 1.87 | 2.81 |
| griewangks with noise $*10^{-2}$ | -4.72 | -6.90 | -7.43 | -8.38 | -9.13 | -9.59 | -10.0 | -10.40 | -10.47 | -10.48 | -10.49 | -10.42 | -10.40 |
| Stochastic function | -6.22 | -7.00 | -7.47 | -7.90 | -8.03 | -8.15 | -8.20 | -8.15 | -8.09 | -7.95 | -7.76 | -7.52 | 12.73 |
| Mean of iteration numbers what algorithm performed before termination | | | | | | | | | | | | | |
| dropwave | 125 | 136 | 138 | 144 | 135 | 112 | 102 | 100 | 98 | 90 | 83 | 71 | 60 |
| griewangks | 81 | 79 | 79 | 78 | 78 | 79 | 80 | 80 | 83 | 85 | 88 | 92 | 94 |
| griewangks with noise | 88 | 92 | 89 | 89 | 89 | 88 | 90 | 90 | 91 | 90 | 88 | 89 | 92 |
| Stochastic function | 97 | 95 | 93 | 94 | 89 | 88 | 84 | 82 | 77 | 75 | 70 | 67 | 62 |
| Mean of iteration numbers when the g_best were found | | | | | | | | | | | | | |
| dropwave | 109 | 93 | 94 | 98 | 89 | 65 | 54 | 51 | 49 | 41 | 34 | 22 | 11 |
| griewangks $*10^{-1}$ | 63 | 62 | 61 | 61 | 61 | 60 | 60 | 61 | 60 | 60 | 57 | 55 | 54 |
| griewangks with noise | 40 | 43 | 40 | 40 | 40 | 39 | 41 | 41 | 42 | 42 | 39 | 40 | 44 |
| Stochastic function | 141 | 140 | 139 | 140 | 136 | 134 | 132 | 130 | 124 | 123 | 118 | 115 | 110 |

Using our method, PSO finds better solution, i.e. the objective value of the best individual and the mean of all objective values in the population is decreased while the number of iterations until the final solution found is decreased also. It yields stronger convergence during the iterations of the algorithm, thus the novel method increases the efficiency of PSO. Obviously the proper setup for the parameters of the PSO and the weight of the gradient is highly problem-dependent, however, during our tests, we found 0.7 as a generally applicable weight for the gradient, and 10 percent of the domain for $\sigma$, which setting in most cases improves the efficiency of PSO. We propose a simple fine-tuning technique in the following to setup the parameters of the algorithm.

1. Set all parameters to zero, i.e. $c_0 = c_1 = c_2 = c_3$.
2. Tune $c_3$ according to the learning method of classic gradient methods, i.e. increase $c_3$ gradually, if oscillation occurs, divide it by 10. Find a stable setting.
3. Set the momentum, i.e. $c_0$, which is typically 0.1 or 0.2 in the literature. Increase it gradually, until some improvement is achieved. Find a stable setting.
4. Tune $c_1$, i.e. increase it gradually until some improvement is achieved.
5. Finally, set $c_2 = 1.25 - c_3$

These technique propose a reliable method for tuning the parameters, however, our tests showed clearly that $c_3 = 0.7$ is a generally good choice, and with $c_1 = 0.5$ and $c_0 = 0.6$ the algorithm operates stable.

### III. STOCHASTIC OPTIMIZATION OF MULTI-ECHELON SUPPLY CHAIN MODEL

Most of the multi-echelon supply chain optimization and analysis are mainly based on analytical approach. Simulation however provides a very good alternative, because it can model real life situations with accuracy, more flexible in terms of input parameters and therefore it is more easy to use in decision support. The simulation results can be analyzed with various statistical methods and numerical optimization algorithms. To analyze complex, especially multi-echelon systems, multi-level simulation models can be used, where the results of optimized high level model feeds into the lower level more detailed models.

The simulation-based approach was published only in the last decade. Jung et al. (Junga, Blaua, Peknya, Reklaitisa & Eversdyk 2004) make a Monte Carlo based sampling from real data, and apply a simulation–optimization framework while looking for managing uncertainty. They use a gradient-based search algorithm, while authors in (Köchel & Nieländer 2005) discuss how to use simulation to describe a five-level inventory system, and optimize this model by genetic algorithm. Schwartz et al. (Schwartz, Wang & Rivera 2006) demonstrate the internal model control (IMC) and model predictive control (MPC) algorithms to manage inventory in uncertain production inventory and multi-echelon supply/demand networks. A complex instance of inventory model can be found in (Hayyaa, Bagchib, Kimc & Sun 2008), where orders cross in time considering various distributions for the lead time. Sakaguchi in (Sakaguchi 2009) investigates the dynamic inventory model in which demands are discrete and varying period by period

The aim of our research is to create a Monte-Carlo simulator which uses probability distributions based on material usage data posted in the logistic module of an enterprise resource planning (ERP) system. The main objective of this development was to build a simulator that can use simple building blocks to construct models of complex supply chain networks. Supply chains processes can be simulated using these modular models, where parameters of Key Performance Indicators are analyzed by sensitivity analysis. The developed SIMWARE simulator can be used as a verification tool to analyze and evaluate inventory control strategies (Király, Belvárdi & Abonyi 2011, Király, Varga, Belvárdi, Gyozsán & Abonyi 2012). The simulation of "actual" inventory controlling strategies provides the most important key performance indicators KPI-s of these strategies. On the other hand this simulator can be used for optimization to determine the optimal values of the key inventory control parameters.

The proposed SIMWARE software provides a framework to analyze the cost structure and optimize inventory control parameters based on cost objectives. With this tool we have minimized the inventory holding cost by changing the parameters of the reordering strategy while keeping the service level at the required value. The simulation of "actual" inventory controlling strategies provides the most important KPI-s of these strategies. On the other hand we can use the simulator as part of optimization and determine the optimal values of the key inventory control parameters. We have minimized the inventory holding cost by changing the parameters of our operational space while keeping the service level at the required value.

### A. Inventory model of a single warehouse

The modular model of the supply chain is based on the following classic model of inventory control. This session gives a summary of the most important parameters of this model. In Figure 3, $Q$ is the theoretical demand over cycle time $T$ and this is the *Order Quantity*; $R$ is the *Reorder point*, which is the maximum demand can be satisfied during the replenishment lead time ($L$). The *Cycle time* ($T$) is the time between two purchase orders. The *Order Quantity* is $Q$, where $Q = \bar{d} \cdot T$. This is the ordered quantity in a purchase order, and $Q$ is equal to the *Expected demand* and the Maximum *stock level. Maximum stock level* is the stock level necessary to cover the *Expected demand* in period $T$; therefore it has to be the quantity we order. *Lead time*($L$) is the time between the Purchase order and the goods receipt. $\bar{d}_L$ denotes the average demand during the replenishment lead time. $\bar{d}_L = \bar{d} \cdot L$, where $\bar{d}$ is the daily average demand. Using the same logic, $\bar{d}_L$ is a special case; it yields consumption if the service level is 100%. We will use $\bar{d}_L$ to denote the consumption during the paper. *Reorder point* is the stock level when the next purchase order has to be issued. It is used for materials where the inventory control is based on actual stock levels.

$S$ is the *Safety stock*; this is needed if the demand is higher than the expected (line *d*). In an ideal case $R$ equals to total of safety stock and average demand over lead time: $R = \bar{d}_L + S$, where $S$ is the *Safety stock* which is defined to cover the stochastic demand changes. For a given *Service Level* this is the maximum demand can be satisfied over the Lead time.

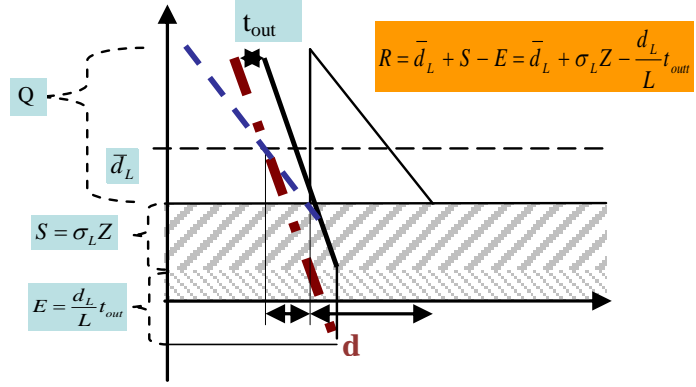

Figure 3: The classic model of inventory control.

Assuming constant demand pattern over the cycle time, Average Stock (*K*) can be calculated as a weighted average of stock levels over the cycle time:

$$K = \frac{Q}{2} + S \tag{15}$$

Service Level (*SL*) is the ratio of the satisfied and the total demand (in general this is the mean of a probability distribution), or in other words it is the difference between the 100% and the ration of unsatisfied demand:

$$SL = 100 - 100\frac{(d_L - R)}{Q} \tag{16}$$

We assume that all demand is satisfied from stock until stock exists. When we reach stock level *R* the demand over the lead time ($\bar{d}_L$) will be satisfied up to *R*. Consequently if $\bar{d}_L < R$, we are getting a stock out situation and there will be unsatisfied demand therefore the service level will be lower than 100%. $\bar{d}_L$ is not known and it is a random variable. The probability of a certain demand level is $P(\bar{d}_L)$. Based on this, the service level is formed as shown in the next equation:

$$SL = 100 - 100\frac{\int_{d_L}^{d_{max}} P(d_L)(d_L - R)d_L}{Q} \tag{17}$$

where $\bar{d}_L$ is continuous random variable, and $\bar{d}_{max}$ is the maximum demand over Lead time. Calculation of SL in practice is simple since probabilities are calculated as frequencies of discrete events and integral is replaced by simple summation of the differences of satisfied and unsatisfied demands.

Based on our experience in analyzing actual supply chain systems we discovered that the probability functions of material flow and demand are different from the theoretical functions (see Figure 4 that shows the distribution function of an actual material consumption compared to the normal distribution used in most of the analytical methodologies). This difference makes difference between the theoretical (calculated) and the actual inventory movements, therefore it makes sense using a stochastic simulation approach based on "empirical" distribution functions.

Inventory movements can be modeled much better using stochastic differential equations than modeling based on the theoretical assumption that movements are following normal distribution. We propose the following model:

$$x_{L_{i+1}} = x_{L_i} - W_i + u(x, R, t_u) \qquad (18)$$

Where $x_i$ is stock level on the $i^{th}$ week, $W_i$ is a stochastic process to represent consumption. This stochastic process is based on the empirical cumulative distribution function we described in the previous section. $u$ is the quantity of material received on week $i$, based on purchase orders. Purchase orders are calculated based on the actual inventory level ($x$), and the replenishment lead-time ($t_u$).

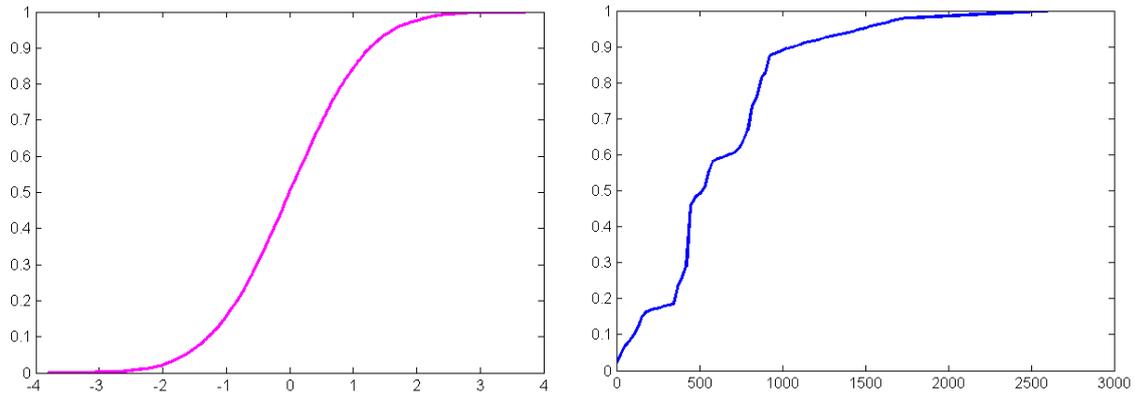

Figure 4.: The theoretical cumulative distribution function (top) and the actual cumulative distribution function for a raw material based on its consumption data (bottom)

### B. *Inventory model of a supply chain*

The main objective of this work is to develop the classical PSO algorithm applying the gradients of the objective function as it was shown in Chapter II. Figure 5 shows the supply chain, i.e. the structure of the analyzed 2-level system. The investigated case study is a two-level inventory system in which there is a central warehouse from only one local warehouse can order ($y_{12}$). Only one product is stored in these warehouses. The customers can buy from the local ($y_1$) and also directly to the central warehouses ($y_2$). To simulate the customers purchase behavior two normal distribution functions are applied as $y_1$ and $y_2$. The mean value is 60 and 50 in the applied distribution functions, while the variance is 15 and 10 respectively. The mean value represents that 60 units of products are averagely consumed in one week from the central warehouse. The variance represents the uncertainty of the mean value, in one week there can be more customers than the mean value, in the other can be less as in real life.

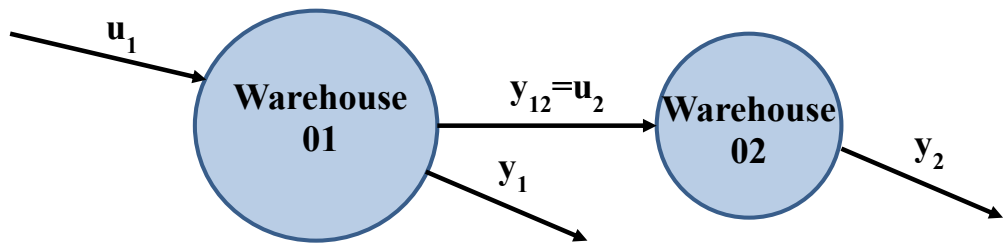

Figure 5: The analyzed 2-level system

The analyzed time period is 50 weeks ($n_{week}=50$). MC simulations are performed to simulate the stochastic behavior of the analyzed warehouses. After the simulations the average properties of the warehouses are calculated. The service levels of both warehouses are determined, they are 0.98 and 0.95. The main objective in the chosen case study is to find the optimal reorder points for both warehouses when the applied objective cost function is at the lowest:

$$f(\text{reorder points}) = \sum_{i=1}^{n_{week}} \left[ \frac{\sum_{j=1}^{n_{MC}} (x_{j,i,1})}{n_{MC}} \cdot HC_1 + \frac{\sum_{j=1}^{n_{MC}} (x_{j,i,2})}{n_{MC}} \cdot HC_2 \right], \qquad (19)$$

where i represents the actual week, j is the actual MC simulation, $x_{j,i,1}$ means the inventory level in the central warehouse at $i^{th}$ week in the $j^{th}$ MC simulation, and $HC_1$ represents the holding cost in the central warehouse. The fluctuations in the average inventory levels after ten MC simulations ($n_{MC}=10$) are shown in Figure 6 when the reorder points are 500 and 200 respectively. It can be seen that before optimization the average inventory of the Warehouse 01 is depleted many times and the minimal stock in Warehouse 02 also reaches zero at the $25^{th}$ week. At the initial reorder points the actual service levels are below the desired values in both of the warehouses (0.60 and 0.89).

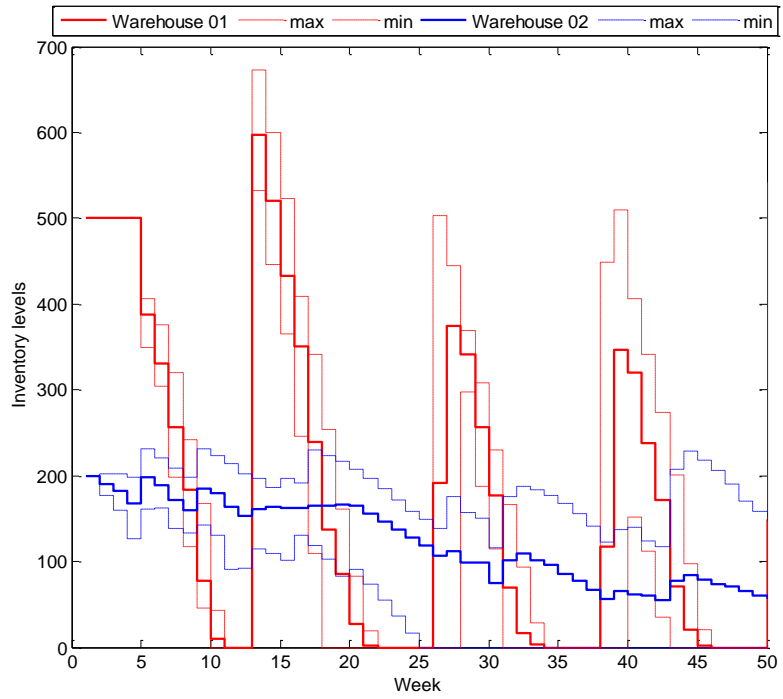

Figure 6: Average inventory levels before optimization

(reorder points 500 and 200 respectively)

### C. Optimization results

The improved PSO algorithm is applied to modify the reorder points due to the value of the objective function and finally to find the global optima. Since the value of the objective function is very high the chosen weight of the gradient for the search is $10^{-3}$. After the optimization process the reorder points is changed to 1031 and 100. Due to this modification the service levels are much better than at the initial state (0.99 and 0.91) and the value of the objective function is $3.42 \cdot 10^5$. The improved PSO finishes the search after 127 generations because in the last 50 generations there was no significant improvement in the value of the objective function. The average inventory levels after optimization can be seen in Figure 7. Due to the optimization the inventory in the central warehouse is not empty in the crucial periods and the minimal stock in the Warehouse 02 is at zero fewer weeks than before the optimization.

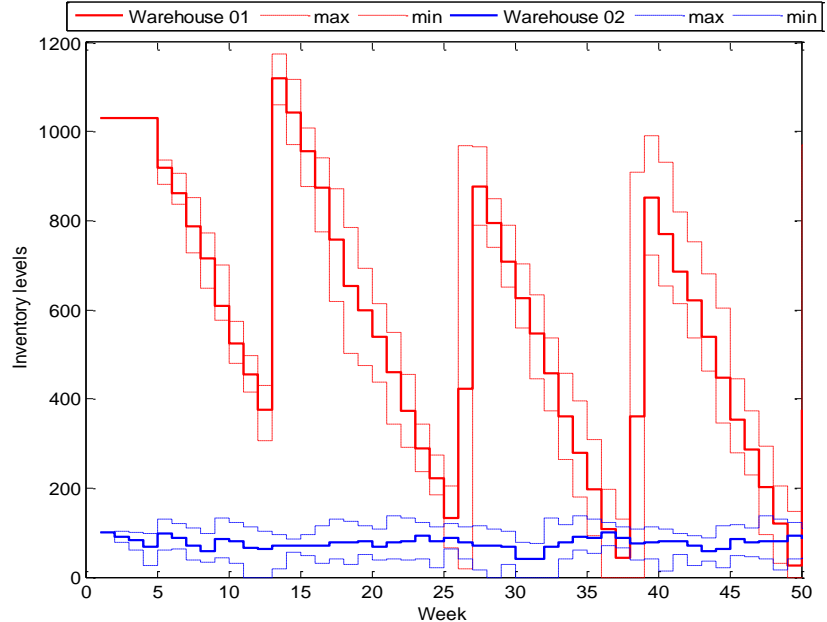

Figure 7: Average inventory levels after optimization

(reorder points 1031 and 100 respectively)

# IV. CONCLUSION

Since supply chain performance impacts the financial performance companies, it is important to optimize and analyze their performances. To support Monte – Carlo analysis of complex supply chains an interactive simulator, SIMWARE has been developed. The stochastic nature of the problem requires effective and robust nonlinear optimization algorithm.

The gradient-free Particle Swarm Optimization (PSO) algorithm is efficient for problems when derivatives do not exist. Application of estimated gradients can boost up the convergence of the PSO. However classical gradient calculation cannot be applied to stochastic and uncertain systems, only robust and local estimates of the gradients can improve converge.

The disadvantage of existing methods of local gradient estimation is the large number of function evaluations required to calculate the gradient of each particles. We developed a more economic, memory based algorithm where numerical approximation of the gradients is based on former function evaluations of the particles. To get local estimates of gradients stored trajectories of particles are weighted based on their distance resulting in weighted least squares regressions. The advantage of this approach is that the size of the region used to calculate the gradients can be controlled by the $\sigma$ parameter.

The performance of the resulted fully informed, regional gradient based PSO is verified by several benchmark problems. The effect of the parameters of the algorithms has been analyzed. The results illustrate the benefits of the incorporation of the regional gradients into the PSO algorithm. Drawback of the method is that it requires careful attention in tuning its parameters ($c_3$ and $\sigma$).

The proposed method is applied in case of multi-echelon system built from two warehouses. We validated our solution by simulating four stochastic input variables. The results illustrate that the developed tool is flexible enough to handle complex situations and straightforward and simple enough to be used for decision support.

ACKNOWLEDGMENT

The financial support of the GOP-1.1.1-11-2011-0045 and TAMOP-4.2.2.A-11/1/KONV-2012-0071 projects are gratefully acknowledged.

BIBLIOGRAPHY

1. Bergh, F. V. D. (2002), An analysis of particle swarm optimizers, PhD thesis, University of Pretoria, South Afrika.
2. Borowska, B. & Nadolski, S. (2009), 'Particle swarm optimization: the gradient correction', *Journal of Applied Computer Science* **17**(2), 7–15.
3. Caloieroa, G., Strozzia, F. & Comenges, J.-M. Z. (2008), 'A supply chain as a series of filters or amplifiers of the bullwhip effect', *International Journal of Production Economics* **114**(2), 631–645.
4. Edwards, A. I., Engelbrecht, A. P. & Franken, N. (2005), Nonlinear mapping using particle swarm optimisation, *in* 'Evolutionary Computation, 2005. The 2005 IEEE Congress on', Vol. 1, IEEE, pp. 306–313.
5. Engelbrecht, A. P., Engelbrecht, A. & Ismail, A. (1999), 'Training product unit neural networks'.
6. Gandomi, A. H. & Alavi, A. H. (2012), 'Krill herd: A new bio-inspired optimization algorithm', *Commun Nonlinear Sci Numer Simulat* **17**, 4831–4845.
7. Graves, S. C. & Willems, S. P. (2008), 'Strategic inventory placement in supply chains: Nonstationary demand', *Manufacturing & Service Operations Management* **10**(2), 278–287.
8. Hayyaa, J. C., Bagchib, U., Kimc, J. G. & Sun, D. (2008), 'On static stochastic order crossover', *International Journal of Production Economics* **114**(1), 404–413.
9. Hu, X. & Eberhart, R. (2002), Solving constrained nonlinear optimization problems with particle swarm optimization, *in* 'Proceedings of the sixth world multiconference on systemics, cybernetics and informatics', Vol. 5, Citeseer, pp. 203–206.
10. Junga, J. Y., Blaua, G., Peknya, J. F., Reklaitisa, G. V. & Eversdyk, D. (2004), 'A simulation based optimization approach to supply chain management under demand uncertainty', *Computers & chemical engineering* **28**(10), 2087–2106.
11. Kennedy, J. & Eberhart, R. (1995), Particle swarm optimization, *in* 'Neural Networks, 1995. Proceedings., IEEE International Conference on', Vol. 4, IEEE, pp. 1942–1948.
12. Király, A., Belvárdi, G. & Abonyi, J. (2011), 'Determining optimal stock level in multi-echelon supply chains', *Hungarian Journal of Industrial Chemistry* **39**(1), 107–112.
13. Király, A., Varga, T., Belvárdi, G., Gyozsán, Z. & Abonyi, J. (2012), Monte carlo simulation based sensitivity analysis of multi-echelon supply chains, *in* 'Factory Automation', Veszprém, Hungary.
14. Köchel, P. & Nieländer, U. (2005), 'Simulation-based optimisation of multi-echelon inventory systems', *International journal of production economics* **93**, 505–513.

15. Makajic-Nikolic, D., Panic, B. & Vujoševic, M. (2004), Bullwhip effect and supply chain modelling and analysis using cpn tools, *in* 'Fifth Workshop and Tutorial on Practical Use of Colored Petri Nets and the CPN Tools'.
16. Mendes, R., Kennedy, J. & Neves, J. (2004), 'The fully informed particle swarm: simpler, maybe better', *Evolutionary Computation, IEEE Transactions on* **8**(3), 204 – 210.
17. Miranda, P. A. & Garrido, R. A. (2004), 'Incorporating inventory control decisions into a strategic distribution network design model with stochastic demand', *Transportation Research Part E: Logistics and Transportation Review* **40**(3), 183–207.
18. Miranda, P. A. & Garrido, R. A. (2009), 'Inventory service-level optimization within distribution network design problem', *International Journal of production economics* **122**(1), 276–285.
19. Nagatania, T. & Helbing, D. (2004), 'Stability analysis and stabilization strategies for linear supply chains', *Physica A: Statistical Mechanics and its Applications* **335**(3), 644–660.
20. Noel, M. M. & Jannett, T. C. (2004), Simulation of a new hybrid particle swarm optimization algorithm, *in* 'System Theory, 2004. Proceedings of the Thirty-Sixth Southeastern Symposium on', IEEE, pp. 150–153.
21. Parsopoulos, K. E. & Vrahatis, M. N. (2002), 'Particle swarm optimization method for constrained optimization problems', *Intelligent Technologies–Theory and Application: New Trends in Intelligent Technologies* **76**, 214–220.
22. Prékopa, A. (2006), 'On the hungarian inventory control model', *European journal of operational research* **171**(3), 894–914.
23. Sakaguchi, M. (2009), 'Inventory model for an inventory system with time-varying demand rate', *International Journal of Production Economics* **122**(1), 269–275.
24. Salomon, R. & Arnold, D. (2009), The evolutionary-gradient-search procedure in theory and practice, *in* R. Chiong, ed., 'Nature-Inspired Algorithms for Optimisation', Vol. 193 of *Studies in Computational Intelligence*, pp. 77–101.
25. Schwartz, J. D., Wang, W. & Rivera, D. E. (2006), 'Simulation-based optimization of process control policies for inventory management in supply chains', *Automatica* **42**(8), 1311–1320.
26. Seo, Y. (2006), 'Controlling general multi-echelon distribution supply chains with improved reorder decision policy utilizing real-time shared stock information', *Computers & Industrial Engineering* **51**(2), 229–246.
27. Sousaa, T., Silvaa, A. & Neves, A. (2004), 'Particle swarm based data mining algorithms for classification tasks', *Parallel Computing* **30**(5), 767–783.
28. Srinivasan, M. & Moon, Y. B. (1999), 'A comprehensive clustering algorithm for strategic analysis of supply chain networks', *Computers & industrial engineering* **36**(3), 615–633.
29. Vaughan, T. S. (2006), 'Lot size effects on process lead time, lead time demand, and safety stock', *International Journal of Production Economics* **100**(1), 1–9.
30. Victoirea, T. A. & Jeyakumar, A. (2004), 'Hybrid pso–sqp for economic dispatch with valve-point effect', *Electric Power Systems Research* **71**(1), 51–59.
31. Wimalajeewa, T. & Jayaweera, S. K. (2008), 'Optimal power scheduling for correlated data fusion in wireless sensor networks via constrained pso', *Wireless Communications, IEEE Transactions on* **7**(9), 3608–3618.
32. Windisch, A., Wappler, S. & Wegener, J. (2007), Applying particle swarm optimization to software testing, *in* 'Proceedings of the 9th annual conference on Genetic and evolutionary computation', ACM, pp. 1121–1128.

33. Yang, X.-S. (2010*a*), 'Firefly algorithm, stochastic test functions and design optimisation', *International Journal of Bio-Inspired Computation* **2**(2), 78–84.
34. Yang, X.-S. (2010*b*), A new metaheuristic bat-inspired algorithm, *in* 'Studies in Computational Intelligence, 2010. Proceedings., Nature Inspired Cooperative Strategies for Optimization (NISCO 2010)', Vol. 4, pp. 1942–1948.
35. Zhang, R., Zhang, W. & Zhang, X. (2009), A new hybrid gradient-based particle swarm optimization algorithm and its applications to control of polarization mode dispersion compensation in optical fiber communication systems, *in* 'Computational Sciences and Optimization, 2009. CSO 2009. International Joint Conference on', Vol. 2, IEEE, pp. 1031–1033.